\documentclass[10pt,twocolumn,letterpaper]{article}
\pdfoutput=1
\usepackage{cvpr}
\usepackage{times}
\usepackage{epsfig}
\usepackage{graphicx}
\usepackage{amsmath}
\usepackage{amssymb}
\usepackage{multirow}
\usepackage{subcaption}
\usepackage{url}

\usepackage[pagebackref=true,breaklinks=true,letterpaper=true,colorlinks,bookmarks=false]{hyperref}

\cvprfinalcopy 

\def\cvprPaperID{288} 

\begin{document}

\title{A Restricted Visual Turing Test for Deep Scene and Event Understanding}

\author{Hang Qi\textsuperscript{\dag\footnotemark[1]}, Tianfu Wu\textsuperscript{\dag\footnotemark[1]}, Mun-Wai Lee\textsuperscript{\ddag}, Song-Chun Zhu\textsuperscript{\dag}
\vspace{2mm}\\
\textsuperscript{\dag}Center for Vision, Cognition, Learning, and Autonomy\\ University of California, Los Angeles\\
{\tt\small hangqi@cs.ucla.edu}, {\tt\small \{tfwu, szchu\}@stat.ucla.edu}
\vspace{2mm}\\
\textsuperscript{\ddag}Intelligent Automation, Inc\\
{\tt\small mlee@i-a-i.com}
\vspace{-0.5cm}
}

\maketitle

\begin{abstract}
    This paper presents a restricted visual Turing test (VTT) for story-line based deep understanding in long-term and multi-camera captured videos. Given a set of videos of a scene (such as a multi-room office, a garden, and a parking lot.) and a sequence of story-line based queries, the task is to provide answers either simply in binary form ``true/false" (to a polar query) or in an accurate natural language description (to a non-polar query).  
    Queries, polar or non-polar, consist of view-based queries which can be answered from a particular camera view and scene-centered queries which involves joint inference across different cameras. 
    The story lines are collected to cover spatial, temporal and causal understanding of input videos. 
    The data and queries distinguish our VTT from recently proposed visual question answering in images and video captioning. 
    A vision system is proposed to perform joint video and query parsing which integrates different vision modules, a knowledge base and a query engine. 
    The system provides unified interfaces for different modules so that individual modules can be reconfigured to test a new method. 
    We provide a benchmark dataset and a toolkit for ontology guided story-line query generation which consists of about 93.5 hours videos captured in four different locations and 3,426 queries split into 127 story lines .
    We also provide a baseline implementation and result analyses. 
\end{abstract}

\renewcommand{\thefootnote}{\fnsymbol{footnote}}
\footnotetext[1]{Contribute equally to this work.}
\renewcommand{\thefootnote}{\arabic{footnote}}

\vspace{-2mm}
\section{Introduction}
\vspace{-1mm}
\subsection{Motivation and Objective}
\vspace{-2mm}
\begin{figure}[htbp]
    \centering
    \includegraphics[width=\linewidth]{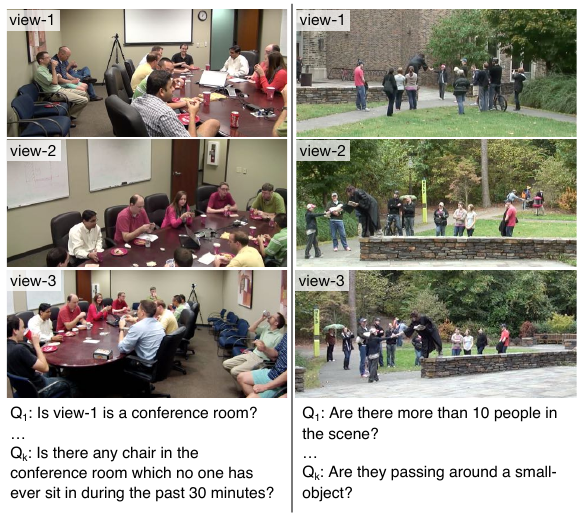}
    \vspace{-5mm}
    \caption{Illustation of depth and complexity of the proposed VTT in deep scene and event understanding, which focuses on a largely unexplored task in computer vision -- joint spatial, temporal and causal understanding of scene and event in multi-camera videos. See text for details. }
    \label{fig:motivation} \vspace{-4mm}
\end{figure}

During the past decades, we have seen tremendous progress in individual vision modules such as image classification~\cite{BOW,PMK,SPM,Tangram} and object detection~\cite{DPM,AOT,grammar,RCNN,fasterRCNN}, especially after competitions like PASCAL VOC~\cite{pascalvoc} and ImageNet ILSVRC~\cite{ILSVRC15} and the convolutional neural networks~\cite{cnn,cnnImagenet,CNN_HumanPerformance} trained on the ImageNet dataset~\cite{imagenet} were proposed. Those tasks are evaluated based on either classification or detection accuracy, focusing on a coarse level understanding of data.
In the area of natural language and text processing, there have been well-studied text-based question answering (QA). For example, a chatterbot named Eugene Goostman\footnote{\url{https://en.wikipedia.org/wiki/Eugene_Goostman}} was reported as the first computer program which has passed the famed Turing test~\cite{Turing} in an event organized at the University of Reading. The success of text-based QA and the recent achievements of individual vision modules have inspired visual Turing tests (VTT) \cite{geman2015,VTT_Fritz} where image-based questions (so-called visual question answering, VQA) or story-line queries are used to test a computer vision system. VTT has been suggested as a more suitable evaluation framework going beyond measuring the accuracy of labels and bounding boxes. Most existing work on VTT focus on images and emphasize free-form and open-ended Q/A's~\cite{VQA1,vqa2015}. 

In this paper, we are interested in a restricted visual Turing test (VTT) --   story-line based visual query answering in long-term and multi-camera captured videos. Our VTT emphasizes a joint spatial, temporal, and causal understanding
of scenes and events, which are largely unexplored in computer vision. By ``restricted", we mean the queries are designed based on a selected ontology.
Figure~\ref{fig:motivation} shows two examples in our VTT dataset. Consider the question how we shall test whether a computer vision system understands, for example, a conference room. In VQA~\cite{vqa2015}, the input is an image and a ``bag-of-questions" (\eg, is this a conference room?) and the task is to provide a natural language answer (either in a multiple-choice manner or with free-form responses). In our VTT, to understand a conference room, the input consists of multi-camera captured videos and story-line queries covering basic questions (\eg, $Q_1$, for a coarse level understanding) and difficult ones (\eg, $Q_k$) involving spatial, temporal, and causal inference for a deeper understanding. More specifically, to answer $Q_k$ correctly, a computer vision system would need to build a scene-centered representation for the conference room (\ie, put chairs and tables in 3D), to detect, track, re-identify, and parse people coming into the room across cameras, and to understand the concept of sitting in a chair (\ie, the pose of a person and scene-centered spatial relation between a person and a chair), etc. 
If a computer vision system can further unfold the intermediate representation to explicitly show how it derives the answer, it enhances the ``trust'' that we have on the system that it has gain a correct understanding of the scene.

\textbf{Web-scale images vs. long-term and multi-camera captured videos.}
Web-scale images emphasize the breadth that a computer vision system can learn and handle in different applications. Those images are often of album photo styles collected from different image search engines such as Flickr, Google, Bing, and Facebook. This paper focuses on long-term and multi-camera captured videos usually produced by video surveillance, which are also important data sources in the visual big data epic and have important security or 
or law enforcement applications. Furthermore, as the example in Figure~\ref{fig:motivation} shows, mutli-camera videos can facilitate a much deeper understanding of scenes and events. The two types of datasets are complementary, but the latter has not been explored in a QA setting. 

\begin{figure*}
    \centering
    \includegraphics[width=\linewidth]{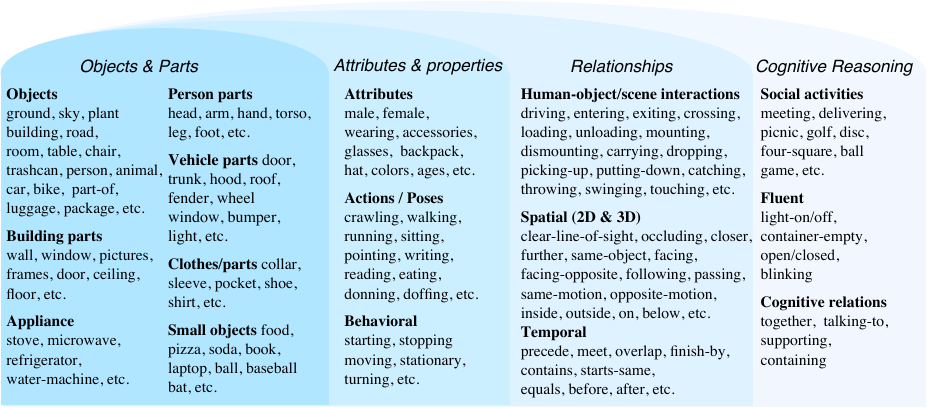}
    \vspace{-0.7cm}
    \caption{The ontology used in the VTT.}
    \label{fig:ontology} \vspace{-0.5cm}
\end{figure*}

\textbf{Free-form and open-ended Q/A's vs. restricted story-line based queries.} Free-form and open-ended Q/A's are usually collected through crowd-sourcing platforms like Amazon Mechanical Turk (MTurk) to achieve diversities. 
However, it is hard to obtain \textit{well-posed} pairs from a massive amount of untrained workers on the Internet.
This is challenging even for simple tasks like image labeling as investigated in the ImageNet dataset~\cite{imagenet} and the Label-Me dataset~\cite{labelme}. For the video datasets in this paper, it is impractical to use MTurk to collect story-line based queries covering long-term temporal ranges and across multi-cameras. Instead, we adopt a selected yet sufficiently expressive ontology (shown in Figure~\ref{fig:ontology}) in generating queries.
Following the statistical principles stated in Geman \etal's Turing test framework~\cite{geman2015}, we design a easy-to-use  toolkit by which several people with certain expertise can create a large number of story lines covering different interesting and important spatial, temporal and, causal aspects in videos with the quality of queries and answers controlled.

\textbf{Quest for an integrated vision system.} Almost all the recent methods proposed for image captioning and VQA are based on the combination of convolutional neural network~\cite{cnn,cnnImagenet} and recurrent neural network like long short-term memory~\cite{lstm}. On the one hand, it is exciting to see much progress have been made in terms of performance. On the other hand, it shows the restricted setting of the tasks in image captioning and VQA. The proposed VTT entails an integrated vision system which cannot be handled by training convolutional and recurrent neural networks directly, to the best of our knowledge. We present a prototype vision system as our baseline implementation which integrates different vision modules (where the state-of-the-art CNN based components can be applied), a knowledge base, and a query engine. 

\begin{figure*}
    \centering
    \includegraphics[width=\linewidth]{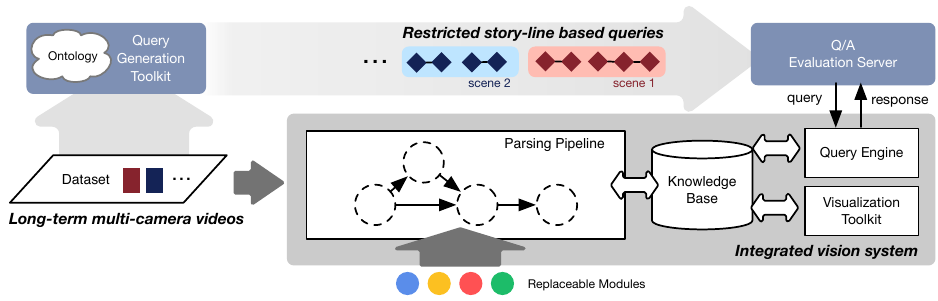}
    \caption{A systematic overview of the proposed VTT. See text for details.}
    \label{fig:introduction} \vspace{-0.3cm}
\end{figure*}

\begin{figure*}[hbtp]
    \centering
    \includegraphics[width=\linewidth]{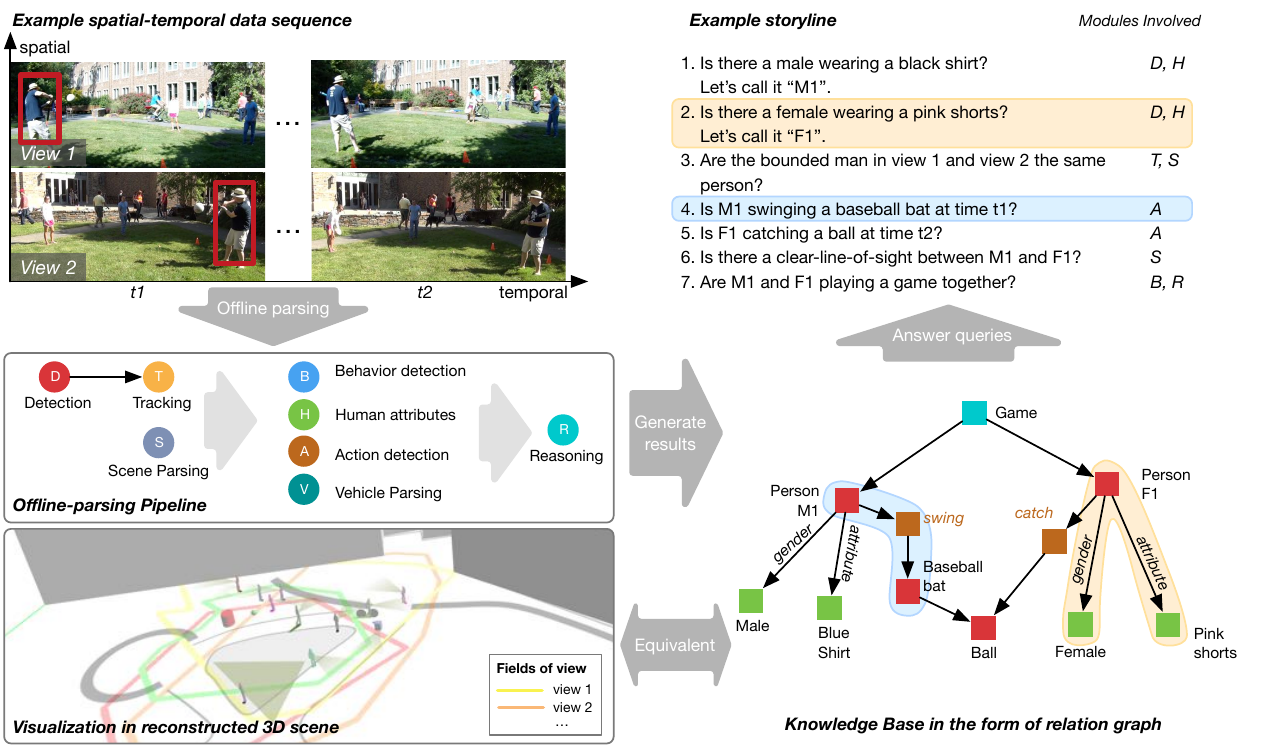}
    \caption{Illustration of our prototype vision system for VTT. \textit{Top-left:} input videos with people playing baseball games. \textit{Middle-Left:} Illustration of the offline parsing pipeline which performs spatial-temporal parsing in the input videos.
        \textit{Bottom-Left:} Visualization of the parsed results. \textit{Bottom-Right:} The knowledge base constructed based on the parsing results in the form of a relation graph. \textit{Top-Right:} Example story line and queries. Graph segments used for answering two of the queries are highlighted. }
    \label{fig:query-example} \vspace{-4mm}
\end{figure*}

\subsection{Overview}
\vspace{-2mm}
Figure~\ref{fig:introduction} illustrates a systematic overview of the proposed VTT which consists of four components: 

\textit{i) Multi-camera video dataset collection:} 
Existing datasets are either focusing on single individual images or
short video sequences with clear action or event boundaries.    
Our multiple-camera video dataset includes a rich set of activities in both indoor and outdoor scenes. Videos are collected by multiple cameras with overlapping field-of-views during the same time window.
A variety types of sensors are used: stationary HD video cameras located on the ground and rooftop, moving cameras mounted on bicycles and automobiles, and infrared cameras. The camera parameters are provided as meta data. The videos capture daily activities of a group of people and different events in a scene which include routine ones (\eg, an ordinary group launch, playing four square soccer game) and abnormal ones (\eg, evacuating from a building during a fire alarm) with large appearance and structural variations exhibited. 

\textit{ii) Ontology guided story-line based query/answer collection:}
We are interested in a selected ontology as listed in Figure~\ref{fig:ontology}. The ontology is sufficiently expressive to represent different aspects of spatial, temporal, and causal understanding in videos from basic level (\eg, identifying objects and parts) to fine-grained level (\eg, does person A have a clear-line-of-sight to person B?). Based on the ontology, we build a toolkit for story-line query generation following the statistical principles stated in~\cite{geman2015}.  Queries organized in multiple story lines are designed to evaluate a computer vision system from basic object detection queries to more complex relationship queries, and further probe the 
system's ability in reasoning from the physical and social perspectives, which entails human-like commonsense reasoning.
Cross-camera referencing queries requires the ability to
integrate visual signals from multiple overlapping sensors.

\textit{iii) Integrated vision system:} 
We build a computer vision system that can be used to
study the organization of modules designed for different tasks
and interactions between them to improve the overall performance.
It is designed with two principles in mind:
first, well-established computer vision tasks shall be 
incorporated so that we can built upon the existing achievements;
second, the modules shall be loosely coupled so that 
it allows user to replace one or more modules with alternatives to study the performance in an integrated environment. 
We define a set of APIs for each individual task
and connect all modules into a pipeline.
After the system has processed the input videos
and saved the results in its knowledge-base,
it fetches queries from the evaluation server one after another at the 
testing time.

\textit{iv) Q/A evaluation server: }
We provide a web service API through which a computer vision system can
interact with the evaluation server over HTTP connections.
The evaluation server iterates through a stream of queries grouped by scenes.
In each scene, queries are further grouped into story lines.
A query is not available to the system
until the previous story lines and all previous queries in the same story line have finished.
The correct answer is provided to the system after each query.
This information can be used by the system to be adaptive with the ability to learn from the provided answers.
The answer can be used to update the previous understanding
such that any conflict has to be resolved and wrong interpretations can be discarded.

Figure~\ref{fig:query-example} shows an example of a full workflow of our system. We have spent more than 30 person-year in total to collect the data and build the whole system. Our prototype system has passed a detailed third-party evaluation involving more than 1,000 queries.
We plan to release the whole system to the computer vision community and organize competition and regular workshop in the near future. 

\section{Related Work and Our Contributions}
\vspace{-2mm}
\label{sec:related-work}
Question answering is the natural way of effective communication between human beings. Integrating computer vision and natural language processing, as well as other modal knowledge, has been a hot topic in the recent development of deeper image and scene understanding. 

\textbf{Visual Turing Test.}
Inspired by the generic Turing test principle in AI~\cite{Turing}, Geman \etal proposed a visual Turing test~\cite{geman2015} for object detection tasks in images which organizes queries into story lines, within which queries are connected and the complexities are increased gradually
-- similar to conversations between human beings.
In a similar spirit, Malinowski and Fritz~\cite{VQA2,VTT_Fritz} proposed a multi-word method to address factual queries of scene images.  
In the dataset and evaluation framework proposed in this paper, 
we adopt similar evaluation structure to \cite{geman2015}, but focus on a more complex scenario
which features videos and overlapping cameras to facilitate a broader
scope of vision tasks.

\textbf{Image Description and Visual Question Answering.}
To go beyond labels and bounding boxes, image tagging~\cite{imageTagging1}, image captioning \cite{imageCaptioning,imageCaptioning1,mRNNCaptioning}, and video captioning~\cite{videoCaptioning} have been proposed recently.   The state-of-the-art methods have shown, however, a coarse level understanding of an image (\ie, labels and bounding boxes of appeared objects)  together with natural language $n$-gram statistics suffices to generate reasonable captions.
Microsoft COCO~\cite{coco2014} provides descriptions or captions for images.
Question answering focuses on specific contents on the image and 
evaluate the system's abilities using human generated question.
Unlike the image description task where a generated sentence is consider
correct as long as it describes the dominant objects and activities in the image,
human generated questions can ask all details and even hidden knowledge 
that require deduction.
In such scenario, a pre-trained end-to-end system may not necessarily perform well
as the question space is too large to be covered by training data.
IQA~\cite{iqa2015} converts image descriptions into Q/A pairs. 
VQA~\cite{vqa2015} evaluates in a free-formed and open-ended questions about images, where the question-answer pairs are given by human annotators.
Although it encourages participants to pursuit a deep and specific understanding
about the image,
it only focuses on the content of the image and does not address
many other fundamental aspects of computer vision like 3D scene parsing,
camera registration, etc.
Moreover, actions are not static concepts, temporal information are largely missing in images.


\noindent \textbf{Our Contributions:}  This paper makes two main contribution to deep scene and event understanding: 
\begin{itemize}
    \item [i)] It presents a new visual Turing test benchmark consisting of a long-term and multi-camera captured video dataset and a large number of ontology-guided story-line based queries. 
    \item [ii)] It presents a prototype integrated vision system consisting of a well-designed architecture, various vision modules, a knowledge base, and a query engine. 
\end{itemize}


\section{Dataset}
\label{sec:dataset}
\vspace{-2mm}

In this section, we introduce the video dataset we collected for the VTT.
In our dataset, we organize data by multiple independent scenes. Each scene consists of video footage from eight to twelve cameras with overlapping fields of view during the same time period.
By now, we have a total number of 14 collections captured at 4 different locations:
two indoor (an office and an auditorium) and two outdoor (a parking lot and a garden).
Table~\ref{tbl:dataset-summary} gives a summary of the data collections.

\begin{table}
    \scriptsize
    \begin{center}
        \resizebox{1.0\hsize}{!}{
            \begin{tabular}{|l|l|l|l|l|}
                \hline
                Collection & Type & Cameras & Event & Length \\
                &  & (Moving)  & duration & hh:mm:ss  \\
                \hline
                \hline
                Office 1 & Indoor & 9 & 56 min & 8:27:23 \\
                Office 2 & Indoor & 12 & 90 min & 17:35:36 \\
                Auditorium 1 & Indoor & 10 (1) & 15 min & 2:29:50 \\
                Auditorium 2 & Indoor & 11 (1) & 48 min & 8:53:24 \\
                Parking lot 1 & Outdoor & 9 (1) & 15 min & 2:41:24 \\
                Parking lot 2 & Outdoor & 11 (2) & 44 min & 8:15:44 \\
                Parking lot 3 & Outdoor & 9 & 12 min & 2:22:00 \\
                Parking lot 4 & Outdoor & 11 (2) & 47 min & 8:14:42 \\
                Parking lot 5 & Outdoor & 11 (1) & 68 min & 13:15:06 \\
                Parking lot 6 & Outdoor & 11 (1) & 23 min & 4:27:44 \\
                Garden 1 & Outdoor & 7 (1) & 15 min & 1:57:01 \\
                Garden 2 & Outdoor & 10 (2) & 41 min & 6:54:38 \\
                Garden 3 & Outdoor & 8 (1) & 27 min & 3:27:00 \\
                Garden 4 & Outdoor & 8 (2) & 34 min & 4:15:56 \\
                \hline
                \hline
                \multicolumn{3}{|r|}{Total} &  8.9 hours & 93:27:28 \\
                \hline
            \end{tabular}
        }
    \end{center}
    \vspace{-0.5cm}
    \caption{Summary of our VTT dataset.}
    \label{tbl:dataset-summary}
\end{table}

\begin{figure}[htbp]
    \centering
    \begin{subfigure}[b]{0.48\linewidth}
        \includegraphics[width=\textwidth]{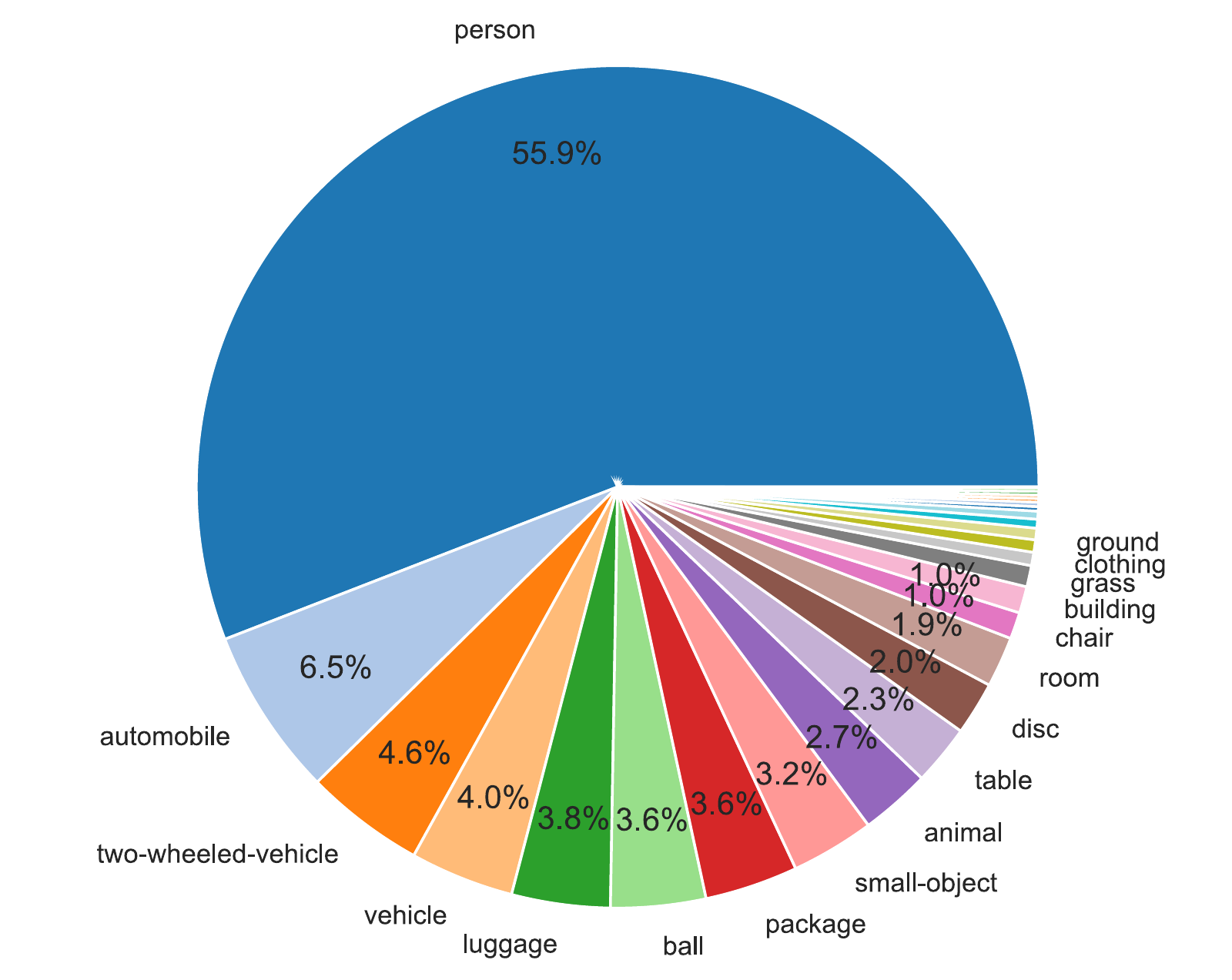}
        \caption{Objects}
        \label{}
    \end{subfigure}
    ~ 
    \begin{subfigure}[b]{0.48\linewidth}
        \includegraphics[width=\textwidth]{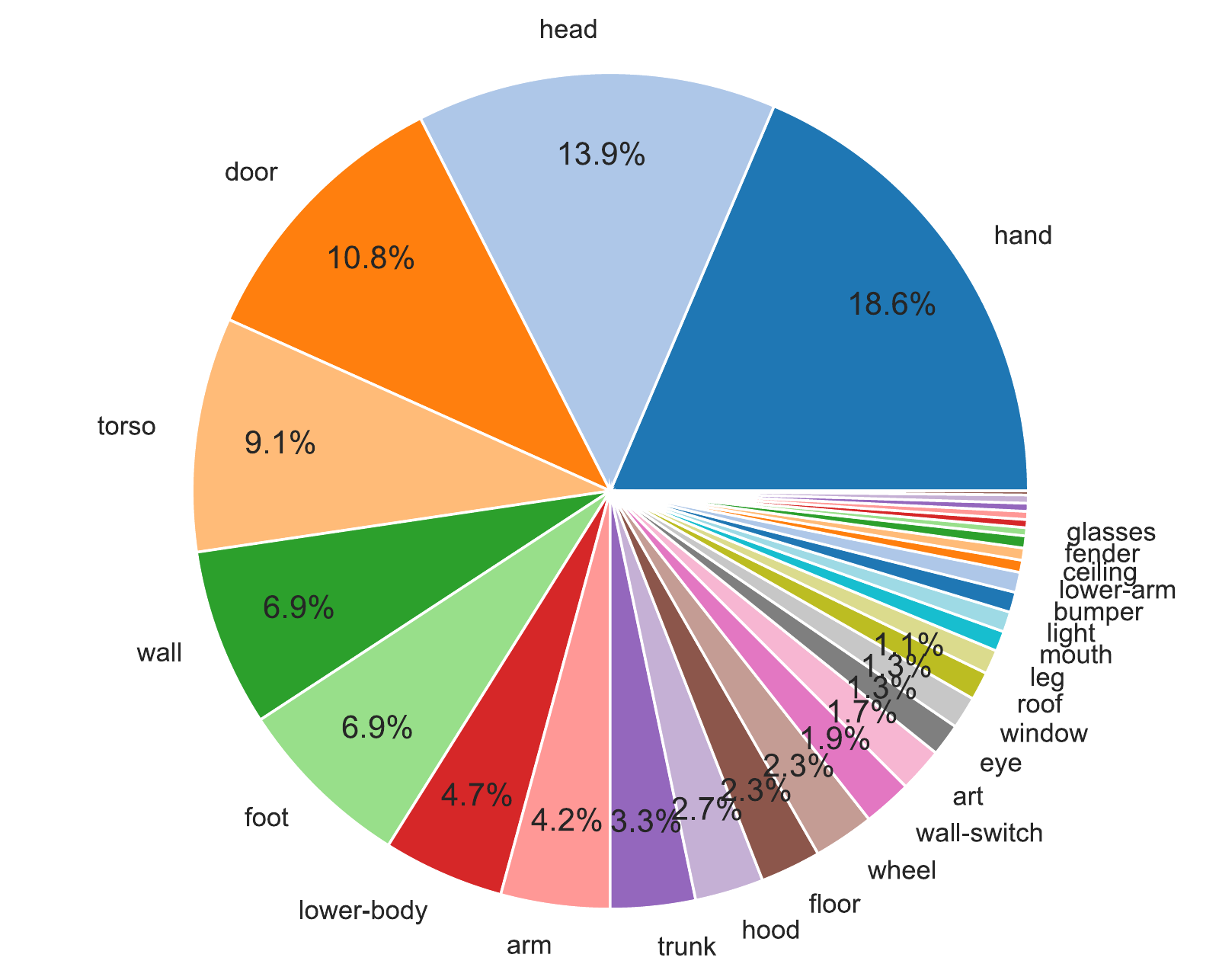}
        \caption{Parts}
        \label{}
    \end{subfigure}
    \\
    \begin{subfigure}[b]{0.48\linewidth}
        \includegraphics[width=\textwidth]{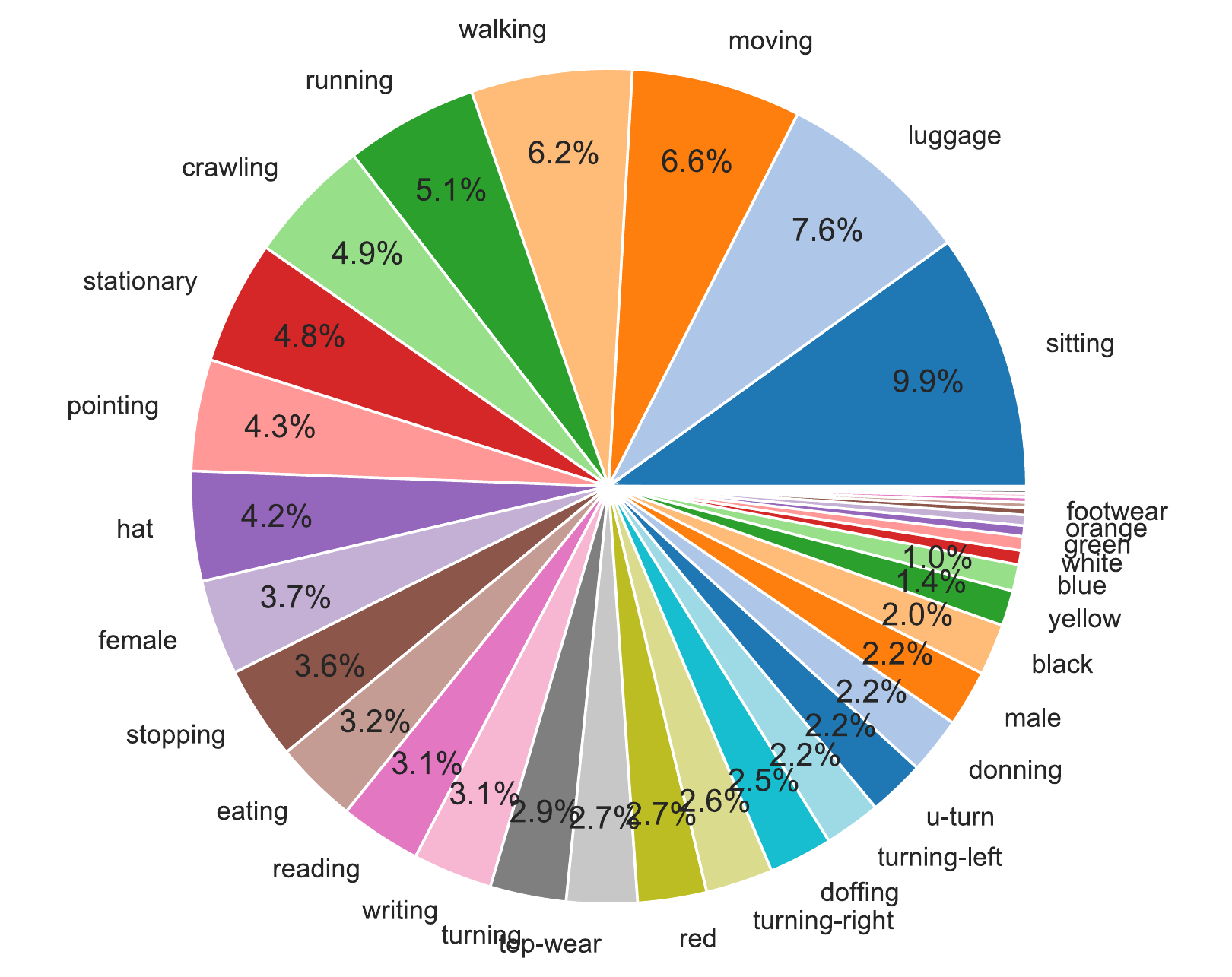}
        \caption{Attributes \& Properties}
        \label{}
    \end{subfigure}
    ~
    \begin{subfigure}[b]{0.48\linewidth}
        \includegraphics[width=\textwidth]{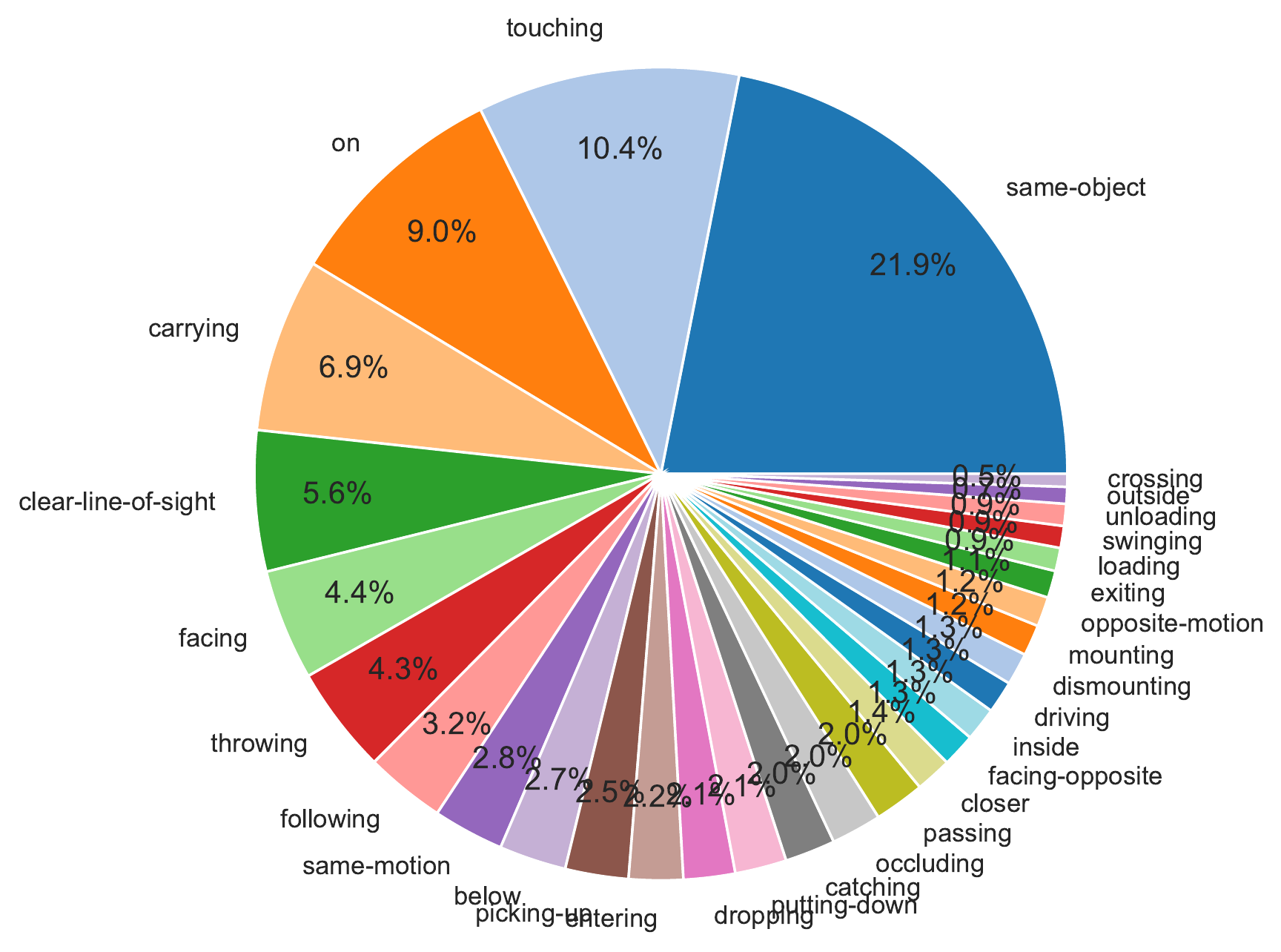}
        \caption{Relationships}
        \label{}
    \end{subfigure}
    \caption{Distribution of predicates}
    \label{fig:predicates}
\end{figure}

Our dataset reflects real-world video surveillance data and poses unique challenges to modern computer vision algorithms:

\textbf{Varied number of entities.}
In our dataset, activities in the scene could involve individuals as well as multiple interacting entities.

\textbf{Rich events and activities.} 
The activities captured in the dataset involves
different degrees of complexities: from the simplest single-person actions 
to the group sport activities which involve as many as dozens of people.

\textbf{Unknown action boundary.}
Unlike existing action or activity dataset where each action data point is
well segmented and each segment only contains one single action,
our dataset consists of multiple video streams.
Actions and activities are not pre-segmented and multiple actions
may happen at the same time.
Such characteristic preserves more information about the spatial context of one action and correlation between multiple actions.

\textbf{Multiple overlapping cameras.}
This requires the system to perform multi-object tracking across multiple cameras with re-identification and 3D geometry reasoning.

\textbf{Varied scales and view points.}
Most of our data are collected in 1920x1080 resolution, 
however, because of the difference in cameras' mounting points,
a person who only occupies a couple of hundred pixels in bird's-eye views may occlude the entire view frame when he or she stands very close
to a ground camera.

\textbf{Illumination variation.}
Areas covered by different cameras have different illumination conditions:
some areas are covered by dark shadows whereas
some other areas have heavy reflection.

\textbf{Infrared cameras and moving cameras.}
Apart from regular RGB signals,
our dataset provides infrared videos as a supplementary.
Moving cameras (\ie, cameras mounted on moving objects) also provide additional challenges to the dataset
and reveal more spatial structure of the scene.

\textbf{The complexity of our VTT dataset.} To demonstrate the difficulties of our dataset, we conduct
a set of experiments on a typical subset of data
using the state-of-the-art object detection models~\cite{fasterRCNN} and multiple-object tracking methods~\cite{pirsiavash2011globally}. 
A summary of the data and results are shown in Table~\ref{tbl:detection-tracking}.

\begin{table}
    \scriptsize
    \begin{center}
        \resizebox{1.0\hsize}{!}{
            \begin{tabular}{|l|l|l|l|l|}
                \hline
                Dataset & Fashion & Sport & Evacuation & Jeep \\
                \hline
                \hline
                Cameras  & 4 & 4 & 4 & 4 \\
                Length (mm:ss) & 4:30 & 1:35 & 3:00 & 3:35 \\
                Frames & 32,962 & 11,798 & 21,830 & 25,907 \\
                \hline
            \end{tabular}
        }
    \end{center}
    
    \begin{center}
        \resizebox{1.0\hsize}{!}{
            \begin{tabular}{|l|l|l|l|l|l|l|l|l|}
                \hline
                Dataset & \multicolumn{4}{c|}{Fashion}                        & \multicolumn{4}{c|}{Sport}\\
                \hline\hline
                Detection & 0.475 & 0.413 & 0.635 & 0.485   & 0.554 & 0.596 & 0.534 & 0.694 \\
                Tracking MOTP
                & 0.683 & 0.674 & 0.692 & 0.694   & 0.728 & 0.727 & 0.716 & 0.739 \\
                Tracking MOTA     & 0.341 & 0.304 & 0.494 & 0.339   & 0.413 & 0.483 & 0.430 & 0.573 \\
                \hline
                \hline
                &\multicolumn{4}{c|}{Evacuation}                         & \multicolumn{4}{c|}{Jeep}\\
                \hline\hline
                Detection & 0.518 & 0.556 & 0.534 & 0.533   & 0.252 & 0.250 & 0.280 & 0.389\\
                Tracking MOTP 
                & 0.698 & 0.692 & 0.720 & 0.651   & 0.680 & 0.651 & 0.689 & 0.696\\
                Tracking MOTA     & 0.389 & -0.241& 0.346 & 0.399   & 0.172 & 0.170 & 0.203 & 0.270\\
                \hline
            \end{tabular}
        }
    \end{center}
    \vspace{-0.5cm}
    \caption{\textit{Top:} Summary of the selected subset of data. \textit{Bottom:} Results from detection and tracking. \textit{For Detection:} 
        AP is calculated as in PASCAL VOC 2012~\cite{pascalvoc}
        based on results by Faster-RCNN~\cite{fasterRCNN}. \textit{For Tracking:} MOTA and MOTP are calculated as in Multiple Object Tracking Benchmark~\cite{MOTChallenge2015} based on results by~\cite{pirsiavash2011globally}.}
    \label{tbl:detection-tracking}
    \vspace{-0.5cm}
\end{table}

\section{Queries}
\label{sec:queries}

A query is a first-order logic sentence (with modification) composed using variables, predicates (as shown in Figure~\ref{fig:ontology}), logical operators ($\wedge, \vee, \neg$), arithmetic operators, and quantifiers ($\exists$ and $\forall$).
The answer to a query is either true or false meaning whether the
fact stated by the sentence holds given the data and the system's state of belief.
The formal language representation eliminates the need of natural language processing
and allows us to focus computer vision problems on a constrained set of predicates.

We evaluate computer vision systems by asking a sequence of queries 
organized into multiple story lines.
Each story line explores a natural event across a period of time
in a way similar to conversations between humans.
At the beginning of a story line, major objects of interest are defined first.
The vision system under evaluation shall indicate whether it detects these objects.
A correct detection establishes a mutual conversation context for consecutive queries,
which ensures the vision system and queries are referring to the same objects 
in later interactions.
When the system fails to detect an object, however,
the evaluation server will skip the queries regarding that object.
Because neither answering these queries correctly nor wrongly
reveals the system's performance in interpreting the designated data.

\textbf{Object definition queries.} To define an object,
specifications of object type, time, and location are three components. Object type is specified by object predicates in the ontology.
A time $t$ is either a view-centric frame number in a particular video
or a scene-centric wall clock time.
A location is either a point $(x, y)$
or a bounding box $(x_1, y_1, x_2, y_2)$ represented by its two diagonal points,
where a point can be specified either in view-centric coordinates (\ie pixels)
or in scene-centric coordinates (\ie latitude-longitude, or coordinates
in a customized reference coordinate system, if defined).
For example, an object definition query regarding a person in the form of first-order logic sentence would look like:
\vspace{-0.2cm}
\[
\exists p \quad \textrm{person}(p; \textrm{time}=t; \textrm{location}=(x_1, y_1, x_2, y_2))
\vspace{-0.2cm}
\]
when the designated location is a bounding box.
Note that the statements made by object definition queries are always true,
as they aim to establish the conversation context.

\textbf{Non-definition queries.}
Non-definition queries in a story line explores a system's
spatial, temporal and causal understanding of events in a scene
regarding the detected objects.
The query space consists of all possible combinations of predicates in the ontology
with the detected objects (and/or objects interacting with the detected ones)
being the arguments.
When expressing complex activities or relationships, 
multiple predicates are typically conjuncted by $\wedge$ to form a query.
For example, suppose $M_1$ and $F_1$ are two detected people confirmed by
object detection queries,
the following query states ``$M_1$ is a male, $F_1$ is a female, and there is a clear line of sight between them at time $t_1$'':
\vspace{-0.1cm}
\[
\small
\textrm{male}(M_1) \wedge \textrm{female}(F_1)
\wedge \textrm{clear-line-of-sight}(M_1, F_1; \textrm{time}=t_1).
\vspace{-0.1cm}
\]
Note that the location is not specified, because once $M_1$ and $F_1$ is 
identified and detected, we assume the vision system can track them over time.
 
Moreover, story lines unfold fine-grained knowledge
about the event in the scene as it goes.
In particular, given the detected objects and established context, 
querying about objects interacting with the detected ones
becomes unambiguous. 
As in the example shown in Figure~\ref{fig:query-example},
even the ball is not specified by any object definition queries
(and actually it is hard to detect the ball even the position is given),
once the two people interacting with the ball  are identified,
it becomes legitimate to ask if ``the female catches a ball at time $t_2$'':
\vspace{-0.2cm}
\[
\exists b\quad \textrm{ball}(b) \wedge \textrm{catching}(F_1, b; \textrm{time}=t_2),
\vspace{-0.2cm}
\]
and if ``the male and female are playing a ball game together over
the period of $t_1$ to $t_2$'':
\vspace{-0.2cm}
\[
\textrm{game}(M_1, F_1; \textrm{time}=(t_1,t_2)).
\vspace{-0.2cm}
\]
Times and locations are specified the same way as in object definition queries
with an extension that a time period $(t_1, t_2)$
can be specified by a starting time and a ending time.

Correctly answering such queries is non-trivial 
as it requires joint cognitive reasoning based on spatial, temporal, and casual information across multiple cameras over a time period.

%

\begin{figure}[htbp]
\footnotesize
\vspace{-1mm}
\begin{verbatim}
  <and>
    <male><entity>p1</entity></male>
    <female><entity>p2</entity></female>
    <clear-line-of-sight>
      <time>t1</time>
      <entity>p1</entity>
      <entity>p2</entity>
    </clear-line-of-sight>
  </and>
\end{verbatim}
\vspace{-5mm}
\caption{An example XML segment of a query in the implementation.
    This segment is equivalent to the statement ``$p_1$ is a male, $p_2$ is a female, and there is a clear line of sight between them at time $t_1$''.}
\vspace{-3mm}
\label{fig:xml-query}
\end{figure}
In non-polar cases, we support three types of questions: ``what'', ``when'', and ``where'', to which the answers are object labels, time intervals, and location polygons, respectively.

Currently, we have created 3,426 queries in the dataset.
Figure~\ref{fig:predicates} shows the distribution of predicates 
in selected categories. Though we try to be unbiased in general,
we do consider some predicates are more common in and important
than others and thus make the distribution non-uniform.
For example, among all occurrence of object predicates,
``person'' takes 55.9\%,
which is reasonable because human activities are our major point of interest.
Meanwhile, we are also building a query generation toolkit on the top of
Vatic~\cite{vatic} for rapid query creation with respect to
the statistical properties discussed by Geman \etal in~\cite{geman2015}.
In the implementation, queries are presented in the form of XML documents
as shown in Figure~\ref{fig:xml-query} for easy parsing.

\section{System}    
\label{sec:system}
\vspace{-2mm}
We designed and implemented a computer vision system
to perform the test as shown in Figure~\ref{fig:introduction}.
It consists of three major parts: an offline parsing pipeline
which decompose the visual perception into multiple sub-tasks,
a knowledge base which stores parsing results
(including entities, properties, and relations between them), and a query engine
which answers queries by searching the knowledge base.
The system also features a flexible architecture and a visualization toolkit.

\subsection{Offline parsing pipeline}
Offline parsing pipeline processes the multiple-view videos.
Each view is first processed by a \textit{single-view parsing pipeline}
where video sequences from multiple cameras are handled independently.
Then \textit{multiple-view fusion} matches tracks from
multiple views, reconciles results from single-view parsing,
and generates scene-based results for answering questions.

To take advantage of achievements in various sub-areas in computer vision, 
we organize a pipeline of modules, each of which focuses on one particular
group of predicates by generating corresponding labels for the input data.
Every module gets access to the original video sequence and
products from previous modules in the pipeline.
The implemented modules are described as follows.
Most components are derived from the state-of-the-art methods at the time we 
developed the system last year.

\textbf{Scene parsing} generates a homography matrix for each sensor by camera calibration and also produces estimated depth map and segmentation label map for
each camera view. The implementation is derived from~\cite{liu2014single}.

\textbf{Object detection}~\cite{song2013,fasterRCNN} processes the video frames and generates
bounding boxes for major objects of interest.

\textbf{Multiple object tracking}~\cite{pirsiavash2011globally} generates tracks for all detected objects.

\textbf{Human attributes}~\cite{parkattributed} classifies appearance attributes of detected human including gender, color of clothes, type of clothes, and accessories (\eg hat, backpack, glasses).


\textbf{Action detection} detects human actions and poses in the scene. The implementation is derived form~\cite{xiaohan2015joint,yao2014animated, wang2011action}.

\textbf{Behavior detection} parses human-human, human-scene, and human-object interactions. 

\textbf{Vehicle parsing}~\cite{wu2015learning,hu2015learning,ramannacar} produces bounding boxes and fluent labels
for specific parts of detected cars (\eg fender, hood, trunk, windows, lights).

\textbf{Multiple-view fusion} merges the tracks and bounding boxes
from multiple views based on appearance and geometry cues.

The middle-left part of Figure~\ref{fig:query-example} shows the dependencies between these modules in the system.

\begin{figure*}[htbp]
\begin{center}
\includegraphics[width=\linewidth]{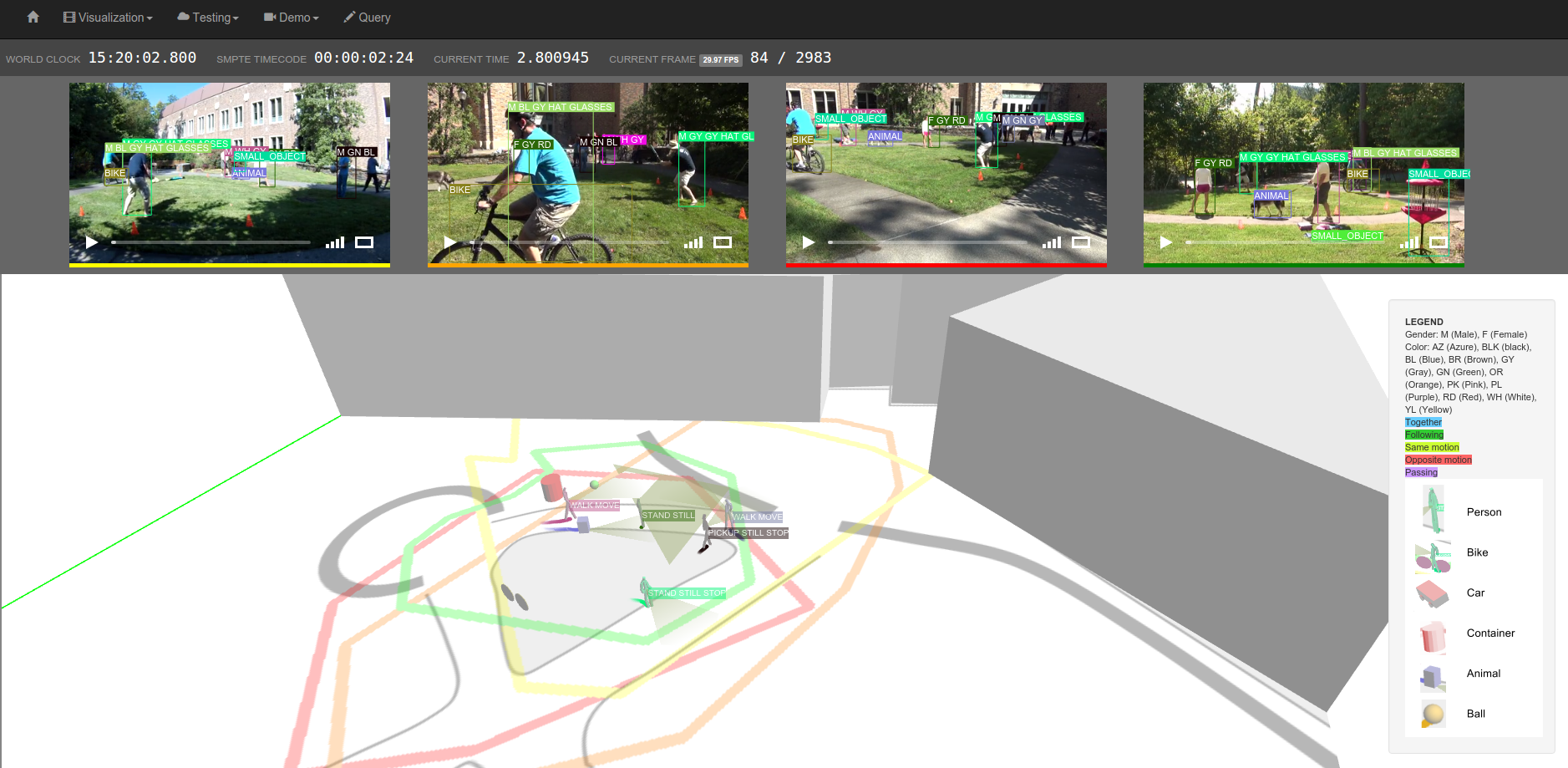}
\caption{Screenshot of the visualization tool. At the top, it shows videos from four different views with detected objects. At the bottom, detected objects are projected into the 3D scene. The videos and the 3D scene share the same playback timeline.}
\label{fig:visualization}
\vspace{-5mm}
\end{center}
\end{figure*}

\subsection{Knowledge base and query answering}
\begin{figure}[htbp]
\centering
\includegraphics[width=\linewidth]{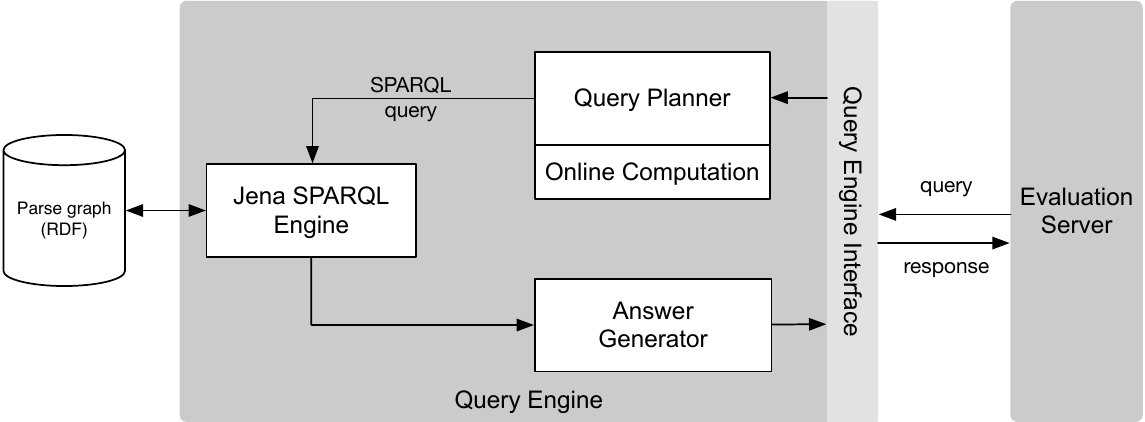}
\caption{Dependencies among single-view parsing tasks.}
\label{fig:query-engine}
\end{figure}

We employ a generic graph-based data model to store knowledge.
The detected objects, actions, attribute labels are all modeled as nodes,
the connections between them are modeled as edges.
In our implementation, 
the parsing results are stored into
Resource Description Framework (RDF) graphs~\cite{rdf},
in the from of triple expressions,
which can be queried by a standard query language SPARQL~\cite{sparql}.
Given that the questions are formal language,
our query engine first parses the query
and transforms the query into a sequence of SPARQL statements.
Apache Jena~\cite{mcbride2002jena} is used to execute these statements and
to return answers derived from the knowledge base.
Figure~\ref{fig:query-engine} shows the architecture of query engine.

In practice, it is infeasible to pre-calculate all possible predicates
and save each individual knowledge segment into the knowledge base.
For example, pre-calculating all ``clear-line-of-sight($x,y$)''
relationships would involve pair-wise combination across all detected humans.
This strategy is obviously inefficient in that the portion of data 
being queried with this predicate is actually sparse.
Alternatively, we designed a \textit{online computation} module
which evaluates binary and trinary relationships only at the testing time
when such predicates appear in a query.

\textbf{Evaluation protocols.}
The computer vision system talks to the evaluation server over HTTP connections.
At the beginning of the evaluation,
the system first acquires an session id from the evaluation server.
Then the system repeatedly request the next available scene, storyline, query in the session from the evaluation server.
In this protocol, the evaluation server maintains the states of evaluation sessions internally and ensures the vision system cannot overwrite the submitted answer to any query.

\subsection{Design Decisions}

The system is architected with two goals bearing in mind: 
first, we want to incorporate existing tasks in computer vision;
second, the architecture shall be flexible enough for replacing a module
with alternatives to pursuit incremental improvements later.
To this end, we defined a set of APIs for each vision task
and connect all the modules using remote procedure calls (RPC).
This enables the system to only focus on the logical connection between
modules and provides the implementation flexibility for individual components.
In practice, we deploy all modules onto different dedicated machines.
Under the RPC interfaces,
computation-intensive algorithms usually utilize GPU and MPI
internally to pursuit faster calculation and data parallelism.
This design allows us to use this system as an experiment platform 
by switching between alternative models and implementations for studying
their effects and contributions to query answering.

To make the system easy to use, we also developed a dashboard with 
visualization tools for rapid development and experiment.
Figure ~\ref{fig:visualization} shows a screenshot of the visualization.

\begin{table*}[htbp]
\begin{center}
        \begin{tabular}{|l|c|c|c|c|c|}
            \hline
            & Office & Parking lot (winter) & Parking lot (fall) & Garden & Auditorium  \\
            \hline
            \hline
            Video length  & 17:35:36 & 8:14:42 & 4:27:44 & 4:15:56 & 8:53:24\\
            \# of cameras & 12 & 12 & 11 & 8 & 11\\
            \# moving cameras & 0 & 2 & 1 & 1 & 2\\
            \# IR cameras & 0 & 1 & 1 & 0 & 1\\
            \# of queries  & 108 & 247 & 236 & 215 & 254\\
            Definition queries &   - &  63 &  71 &  54 &  55 \\
            \hline
            \hline
            Non-definition queries & 108 & 184 & 165 & 161 & 199\\
            Respond rate      & 0.522 & 0.600 & 0.795 & 0.683 & 0.731\\
            \hline
            Accuracy          & 0.785 & 0.615 & 0.626 & 0.586 & 0.684\\
            \hline
        \end{tabular}
    \end{center}
    \vspace{-0.5cm}
    \caption{Performance by data collection.}
    \label{tbl:results-by-data}
\end{table*}

\begin{figure*}[htbp]
    \centering
    \includegraphics[width=\linewidth]{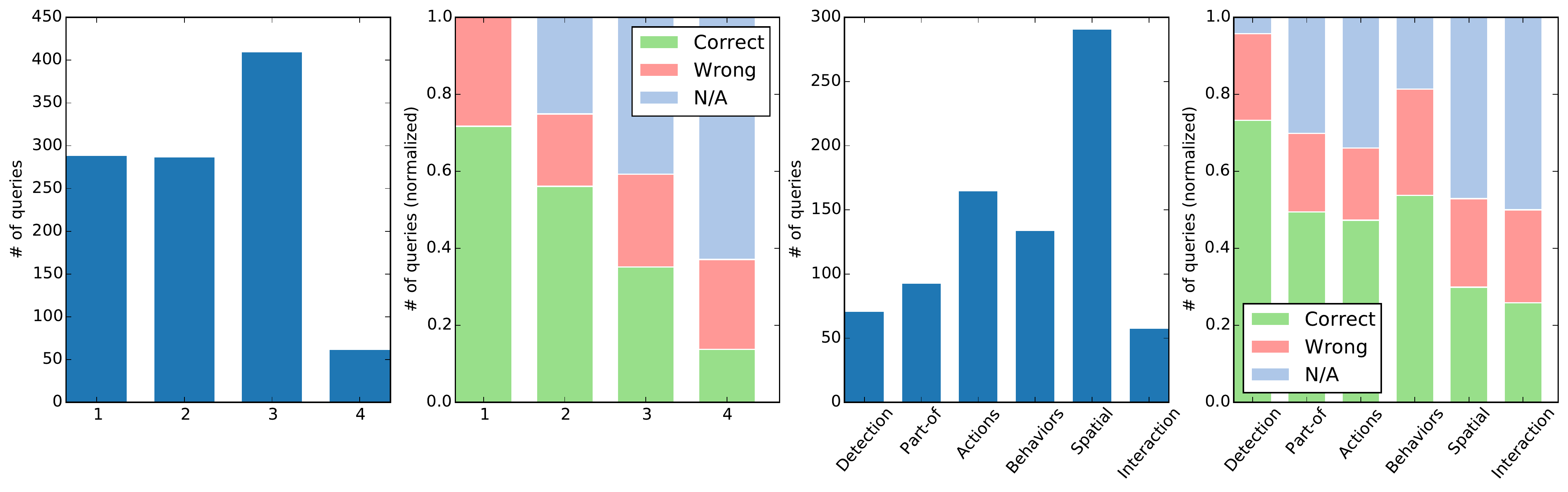}
    \vspace{-8mm}
    \caption{Results breakdown.
    \textit{Left to right}: (1) histogram of unique number of queries by length and (2) accuracies breakdown (object definition queries are included in the calculation; (3) histogram of queries by category and (4) accuracies breakdown.}
    \label{fig:results-breakdown}
\end{figure*}



\section{Evaluation}
\label{sec:evaluation}

Our prototype system has been evaluated by an independent third-party company which collected the datasets and created 1,160 polar queries in a subset of data (see the upper parts in Table~\ref{tbl:results-by-data}). The company was invited to administrate the independent test under the same grant on which we worked.
During the test, the testing data was available to our system two weeks before the story-line query evaluation. We performed the offline parsing within the two weeks by deploying our system on a small cluster consisting of 10 workstations. During the evaluation, our system did not utilize the ground-truth answers received after
each response for consecutive queries. 

Among the 1,160 queries, 243 queries are object definitions,
197 (81\%) of which are successfully detected,
For non-definition queries, we either 
provided binary ``true/false'' answers or
claimed ``unable to respond''
(when our implementation cannot handle or recognize some of the predicates
involved in a query). 
Table~\ref{tbl:results-by-data} shows the accuracy as the ratio of
correctly answered queries to number of the responded non-definition queries.
Note that during the evaluation, for simplicity, the object definition queries are not included in the accuracy calculation,
because they aim to establish mutual knowledge
for consecutive queries in the story line,
which ensures the evaluation server and the system are discussing the same objects.
Therefore, the ground-truth answers to these queries
are actually always ``true''. 
One can obtain an 100\% accuracy in object definition
queries by a trivial method (answering ``true'' at all times) with the risk of 
not discussing the same objects in consecutive queries. Now, we are extending this by generating more object definition queries to which the answers can be ``false"
for evaluating detection performance. 
These queries does not serve to establish conversation context, 
therefore for the story lines starting with an object definition query whose ground-truth answer is false, we randomly sample the predicates and relations to generate the remaining queries. 

Figure~\ref{fig:results-breakdown} further breakdowns the accuracy 
by the number of unique predicates and the category of predicates in a query, respectively.

\textbf{Breakdown by number of predicates.} Most queries have either one, two, or three predicates. This is a natural result of the choice to avoid overcomplicating the queries. As the number of predicates increases, the accuracy of our prototype system decreases, since a wrong prediction in any of the predicates may cause answering the query incorrectly.
The queries with one, two, or three predicates can mostly be explained as follows:

\textit{i) One predicate}: These are queries that deal only with the predicates for the various types of objects (people, car, etc.). Most of these queries (243) are object definition queries; the others (46) deal with counting objects (\eg, ``how many people are in the scene?'').

\textit{ii) Two predicates}: These queries are mostly queries involving unary predicates operating on an object. One predicate is used to define the object (usually person or automobile), and the unary predicate is the second predicate involved.

\textit{iii) Three predicates:} These queries are mostly queries involving binary predicates operating on two objects. Two predicates are used to define the operands, and the binary predicate is the third predicate involved.

\textbf{Breakdown by category.}  When looking at the accuracy by categories, 
our prototype system perform well in classic computer vision tasks (detection, part-of relations, actions, behaviors). However, queries involving spatial reasoning and interactions between human and objects or scene are still challenging
and open to further research.

\section{Discussion and Conclusion}

This paper presented a restricted visual Turing test (VTT) for deeper scene and event understanding in long-term and multi-camera videos. Our VTT emphasizes a joint spatial, temporal and causal understanding by utilizing scene-centered representation and story-line based queries. The dataset and queries distinguish the proposed VTT from the recent proposed visual question answering (VQA). We also presented a prototype integrated vision system which obtained reasonable results in our VTT. 

In our on-going work, we are generating more story-line based queries and setting up a website for holding a VTT competition.
In the proposed competition, we will release the whole system as a playground.
Our system architecture allows a user to substitute one or more modules with their own methods and then run through the VTT to see the improvements.
One of our next steps is to create a publicly available ``vision module market" 
where researchers can evaluate different individual components from the VTT perspective besides the traditional metrics. 





\paragraph{Acknowledgement.} 
This work is supported by
DARPA MSEE FA 8650-11-1-7149 and DARPA SIMPLEX N66001-15-C-4035.
We would like to thank Josh Walters and his colleagues at BAE Systems, the third-party collaborator in the project who administrated the test,
Alexander Grushin and his colleagues at I-A-I for the effort in
testing the system.
We also thank members in the VCLA Lab at UCLA who contributed
perception algorithms in their published work into the baseline test.



{\small
    \bibliographystyle{ieee}
    \bibliography{cvpr16-msee-bib}
}

\end{document}